\pdfoutput=1
\documentclass[11pt,a4paper]{article}
\usepackage[]{acl}
\usepackage{times}
\usepackage{url}
\usepackage{afterpage}
\usepackage{latexsym}
\usepackage{multirow}
\usepackage{wrapfig}
\usepackage{hyperref}
\usepackage[]{graphicx}
\usepackage{subcaption}
\usepackage{csquotes}
\usepackage{amsmath,amssymb}

\DeclareMathOperator{\E}{\mathbb{E}}
\title{Learning to Simplify with Data  Hopelessly Out of Alignment}

\author{Tadashi Nomoto\\
  National Institute of Japanese Literature \\
  10-3 Midori, Tachikawa, Tokyo 190-0014,  Japan\\
  {\tt nomoto@acm.org}\\}

\date{}

\begin{document}
\maketitle
\sloppy
\begin{abstract}
We consider whether it is possible to do text simplification without relying on a \enquote*{parallel}  corpus, one that is made up of sentence-by-sentence alignments of complex and ground truth simple sentences. To this end,  we introduce  a number of concepts, some new and some not,  including  what we call Conjoined Twin Networks, Flip-Flop Auto-Encoders (FFA) and Adversarial Networks (GAN).  A comparison is made between  Jensen-Shannon (JS-GAN) and Wasserstein GAN, to see how they impact performance, with stronger results for the former.   An experiment we conducted with a large dataset derived from Wikipedia found  the solid superiority of Twin Networks equipped with FFA and JS-GAN, over the current best performing system.  Furthermore, we discuss where we stand in a relation to fully supervised methods  in the past literature, and highlight with examples qualitative differences that exist  among simplified sentences generated by supervision-free systems. 
\end{abstract}

\section{Introduction}

What motivated us to embark on the current  work  is the same as one that set 
\cite{DBLP:journals/corr/abs-1710-11041} off on the journey to  unsupervised machine translation.   Any  successful  machine translation requires a large parallel corpus, one containing millions and millions of sentences, each paired with a manual translation. Which naturally gives rise  to a  question: Is it possible to throw away manual translations, which demand so much investment in labor and money? In this work, we aim  to recreate what \cite{DBLP:journals/corr/abs-1710-11041} did with machine translation, for text simplification, building a learning simplification system that works without  ground truth data.

\par \cite{Tomoyuki_Kajiwara2018} recently introduced an approach that might lead to eliminating the need for human intervention for  text simplification.  It involves randomly picking from some corpus \enquote*{hard} sentences,  followed by finding matching \enquote*{easy} sentences that are sufficiently close in meaning.\footnote{
 Whether a sentence is hard or easy can be decided with a reference to a readability metric such as Flesch Reading Ease and  Flesch-Kincaid Grade Level. }
The authors demonstrated that simplification systems trained on a corpus created in this manner performed neck and neck with  those that are trained on human crafted gold standards. 
Which suggests that  a successful simplification system may not require strictly \enquote*{parallel}  data where source sentences are carefully put in line  with targets.   The problem with this approach however, is that the task of matching up sentences  soon becomes daunting with the growth of data.  If we have a corpus of 1 million sentences,  we will have to make $(10^{12} - 10^{6})/2$ comparisons to arrive at a training corpus we need, which leads to an obvious question: Is it possible do away comparison altogether? 

\par This is a question we intend to answer in this work.  We work with data whose source and target sentences are  completely out of alignment,  not even close in meaning. The only condition we impose on our data is just that source sentences are \enquote*{hard} and targets \enquote*{easy}  in terms of readability.  We are interested in whether we can isolate and inject stylistic features of readable sentences into hard sentences, improving the latter's readability.  The approach we take introduces  what we call Conjoined Twin Networks, which seek to bring the latent representation of  a hard sentence closer to  that of an easy sentence via Generative Adversarial Networks (GAN).   We  also take a look at Jensen-Shannon GAN and Wasserstein GAN and discuss what impact they have on text simplification, along with other variants of Twin Networks. 

\section{Related Work}
The last decade has seen an increased effort in the NLP community to take advantage of a fast evolving field of deep learning (DL).  
\cite{wubben:etal:2012:sentence} was a notable exception to this trend. Its goal was to make a simple sentence  by reranking  outputs generated by a statistical translation model (SMT) \cite{koehn:etal:2007:moses},  based on the divergence from the source, as measured by the Levenshtein distance.
\cite{zhang-lapata-2017-sentence}, on the other hand,  tackled  text simplification by  introducing   an LSTM-based sequential model armed with a series of linguistically motivated reinforcement objectives, to ensure that its output has simplicity, fluency and relevance to the source.
\cite{xu-etal-2016-optimizing} developed an approach nearly identical to \cite{wubben:etal:2012:sentence}, composed of an SMT and a reranking mechanism. They differ in that the latter uses   in addition to BLEU, SARI,\footnote{SARI is a divergence metric looking at  how many words are present in the target which do not occur in the source and how many words are absent in the target that are present in the source. There will be more discussion on SARI later. } 
features extracted from  a large paraphrase dictionary,  PPDB \cite{ganitkevitch:etal:2013:ppdb} as objectives.
\cite{surya-etal-2019-unsupervised} is the first attempt  to create  a fully functional simplification system relying solely on unaligned data, the work we discuss extensively later in the paper. The idea is to leverage cross tied multiple translation models along the lines of \cite{DBLP:journals/corr/abs-1711-00043},\footnote{
The authors created a translation system that learns from monolingual texts alone.  This was made possible in part by a particular way the system is structured: it encourages the source it back-translates from the target  to stay semantically close to  the original source sentence.
} together with Generative Adversarial Networks (GAN),  to copy stylistic and lexical features from an easy to hard sentence.  

\section{Conjoined Twin Networks}
\begin{figure}[t]
\begin{center}s\begin{subfigure}[h]{0.3\textwidth}
\includegraphics[height=2.5cm, trim={4.8cm 12cm 14cm 6cm},clip]{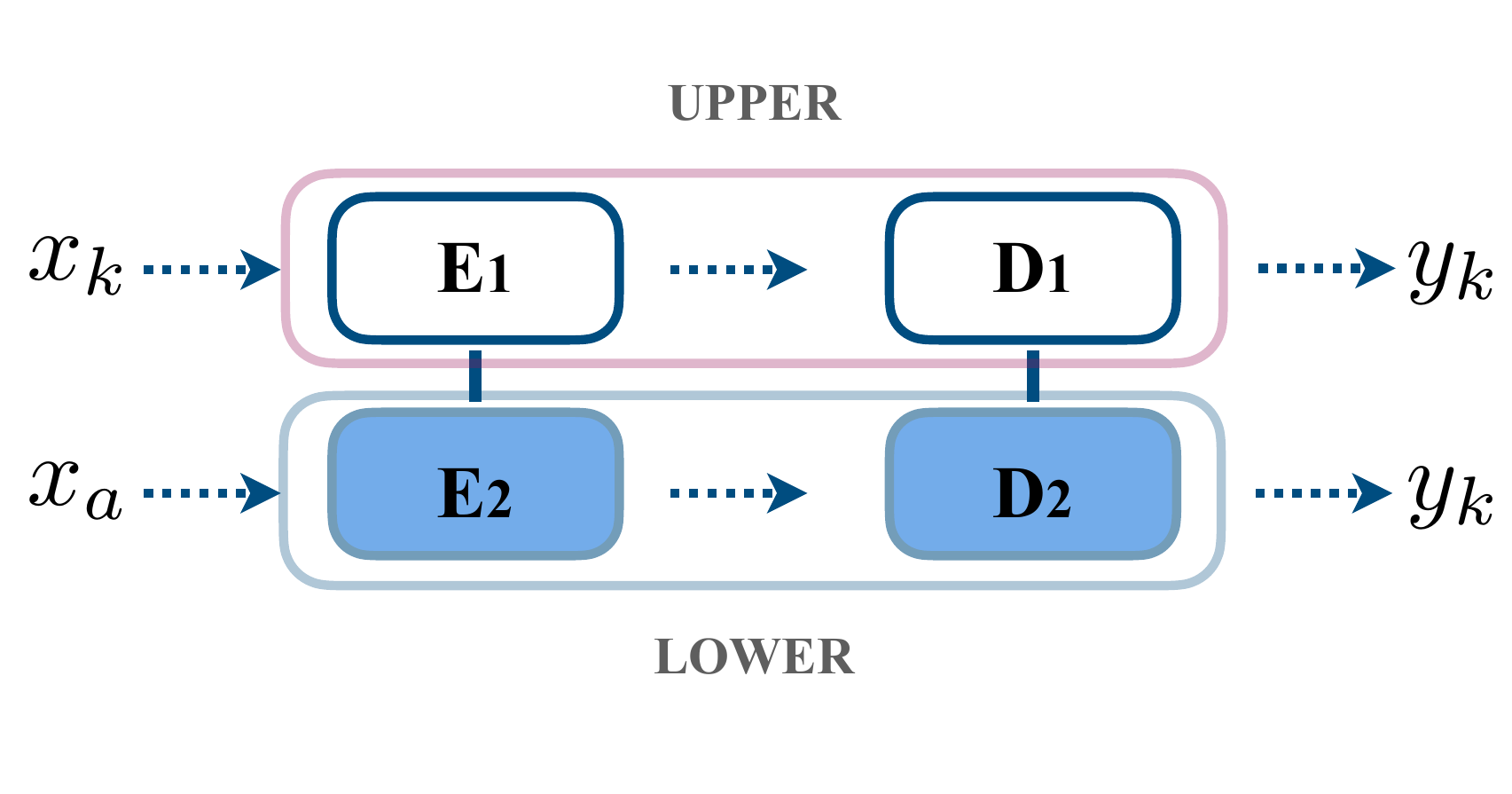}
\caption{Base Twin Networks ($\beta$-TWIN)\label{fig:base_model}} 
\end{subfigure}
\begin{subfigure}[h]{0.35\textwidth}
\includegraphics[height=2.5cm, trim={4.5cm 12cm 10cm 6cm},clip]{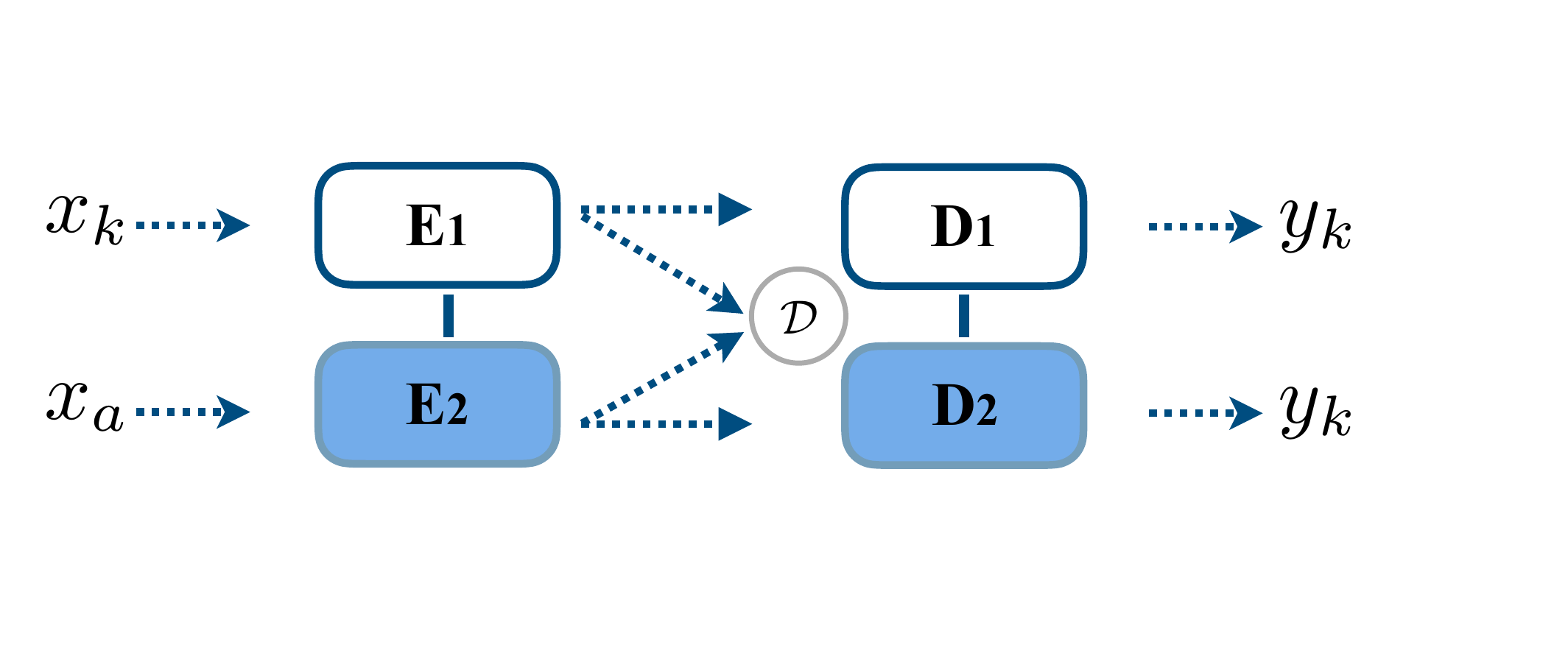}
\caption{GANed  Twin Networks ($\gamma$-TWIN) \label{fig:ganned}}
\end{subfigure} \\
\begin{subfigure}[h]{0.35\textwidth}
\includegraphics[height=2.5cm, trim={3.5cm 12cm 10cm 6cm},clip]{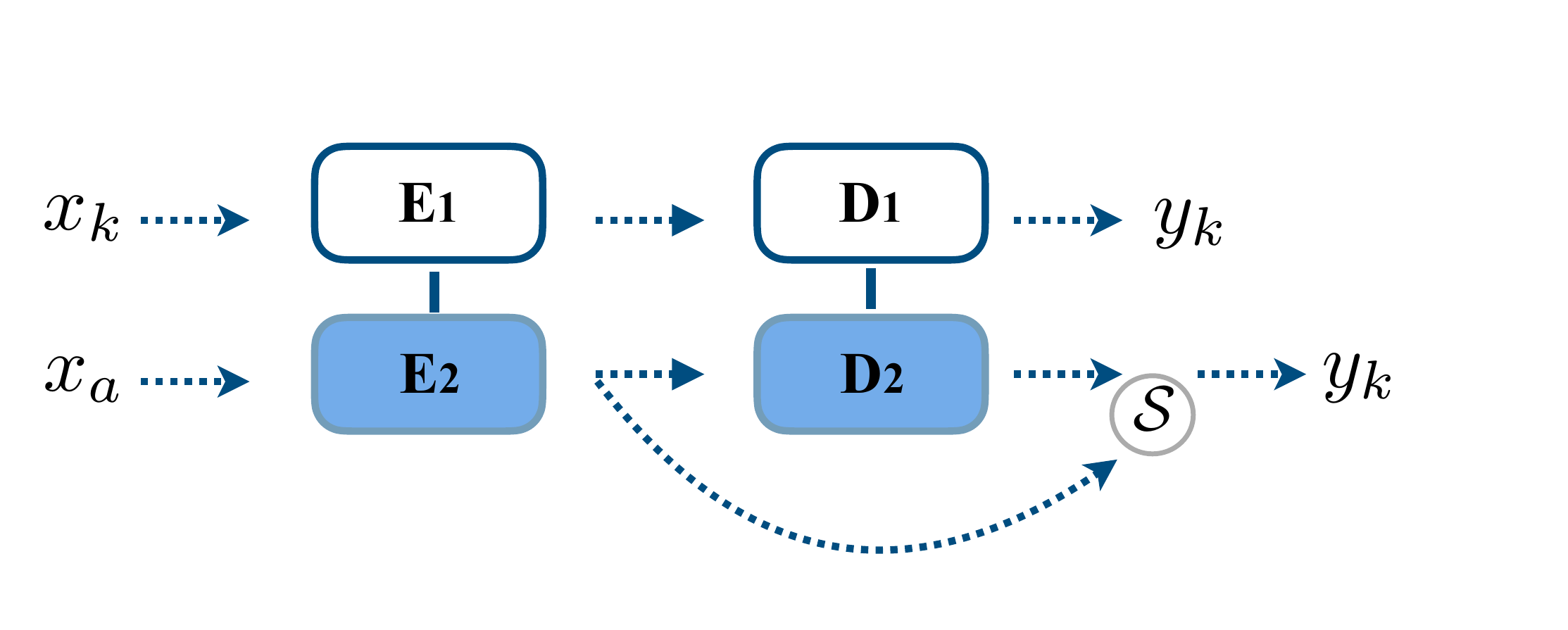}
\caption{ Twin Networks with Similarity Loss ($\sigma$-TWIN) \label{fig:ganned_simloss}}
\end{subfigure} 
\caption{ Conjoined Twin Networks \label{fig:cts}}
\end{center}
\end{figure}
We  consider a particular architecture which consists of  a pair of sequence to sequence networks which are conjoined to each other, i.e.  both the encoder and decoder of each network are shared across the two systems. We call  the architecture  \enquote*{Conjoined Twin  Networks.}  In this work, we look at three variations on the model, shown in Fig.~\ref{fig:cts}. $x_k$  denotes an \enquote*{easy} sentence and $x_a$ a  \enquote*{hard} sentence.  Similarly we will use  $y_k$  and $y_a$ to represent corresponding targets. \enquote*{E} represents  an encoder and \enquote*{D} a decoder.  Fig.~\ref{fig:base_model} is  the simplest of all ($\beta$-TWIN), made up of an auto-encoder  that  works to reconstruct input $x_k$  (call it UPPER)  and a translation model that converts $x_a$  to $y_k$  (call it LOWER).   Note that $y_k = x_k$. %
\par  The goal of Conjoined Twin Networks is to transfer stylistic features such as readability and length from UPPER to LOWER, so that the latter learns to generate a simplified version of $x_a$.  The links between the two $E$s and $D$s mean that they share all weight updates  during the training.

\par Fig.~\ref{fig:ganned} shows a  variant on Twin Networks ($\gamma$-TWIN) which has a discriminator  applied to outputs of  encoders $E_{1}$ and $E_{2}$ or $E_{1}(x_k)$ and $E_{2}(x_a)$.   The aim is to encourage $E_{2}$  to produce a representation similar to one from $E_{1}$.

\par  A Twin Network ($\sigma$-TWIN) in Fig.~\ref{fig:ganned_simloss} represents another  possible way to get $x_a$ and $x_k$ closer. It  relies on  a particular objective  function to  reduce a  difference between   $D_{2}(E_{2}(x_a))$  and  $E_{2}(x_a)$, which  takes the form of  a cosine embedding loss:
\begin{equation}
{\cal S}  =1-\cos(E_{2}(x_a),D_{2}(E_{2}(x_a))). 
\end{equation}

\par We assume that $\beta$-TWIN, $\gamma$-TWIN, and $\sigma$-TWIN  all come equipped with a composite  reconstruction loss: 
\begin{equation}\label{eqn:reconst}
\begin{split}
{\cal L}_{rec} = & \E_{x\sim P_{k}} [ - \log P_{D_1} (y \mid  E_1(x))]  \\
& + \E_{x\sim P_a} [ - \log P_{D_2} (y \mid  E_2(x))]. \\ 
\end{split}
\end{equation}
$P_a$ and $P_k$ are probability distributions of $x_a$ and $x_k$, respectively. 
\subsection{Jensen-Shannon vs. Wasserstein  GAN}
Another focus of this work is  on  whether a choice of objective for GAN has any effect on performance, in particular in the context of text simplification. Much of the prior work in the NLP literature  that incorporates GAN  takes  the Jensen-Shannon divergence as something of a default. It  can be derived from the following by taking the optimal value for discriminator $D$ \cite{Arjovsky2017TowardsPM}. 
\begin{equation}
\begin{split}\label{eqn:js}
\min_G\max_D L(D,G) = &  \E_{x\sim p_r(x)} [ \log D(x) ] \\
& + \E_{x\sim p_g(x)} [ \log (1 - D(x)) ] \\
\end{split}
\end{equation}
$p_r(x)$ represents the prior distribution of real samples and $p_g(x)$ that of fake samples produced by  generator $G$. 
What we should note about JS-GAN is  a unique way its target objective for generator $G$ is set up. Its aim is to minimize: $\E_{z\sim p_z(z)} [ \log (1 - D(G(z)) ]$, or equivalently to maximize $x^\star = \arg\max_x  D(x)$, with $x=G(z)$. $G$ doesn't care about the semantics, or linguistic quality of outputs it generates. All it cares about is whether it can fool $D$ into believing that outputs it gets from $G$ are all genuine and authentic. We expect JS-GAN to be generally exempt from  effects of  linguistic properties, including  semantics,  that are not relevant to determining the readability, thus more adept at finding features  particular to the readability, isolating them from all the other features that may reside in the sentence. 

\par A well-known alternative to JS-GAN is Wasserstein distance, which is based on what is generally known as Earth Mover's Distance or EMD.\footnote{\url{https://en.wikipedia.org/wiki/Earth_mover\%27s_distance}}  EMD is motivated by the idea that the distance or dissimilarity between any two distributions of events or goods can measured by how much work is involved in making one distribution look like the other. 
\begin{table*}
\caption{Discriminator and Generator in JS-GAN  and W-GAN. \label{tbl:train-gans}}
\center \begin{tabular}{ccc}
& \sc discriminator loss & \sc generator loss\\
JS-GAN& $\E [ \log  D(E_1(x))]  +\E[ \log (1 - D(E_2 (x)) ]$  &  $ \E[\log D(E_2(x))]$ \\\
W-GAN&  $ \E[ f(E_1(x)) ]  -  \E [f(E_2 (x))] + \lambda\,{G_p} $ & $ \E[f(E_2 (x)) ] $\\
\end{tabular}
\end{table*}
\begin{table}
\caption{Loss Functions for Twin Networks \label{tbl:twin-loss}}
\hspace{-4mm}\begin{tabular}{l|p{.5\textwidth}}
\hspace{4pt}$\beta$-TWIN & ${\cal L}_{rec}$ \\
\hspace{4pt}$\gamma$-TWIN$_j$ & ${\cal L}_{rec} +  \E[\log D(E_2 (x))] $ \\
\hspace{4pt}$\gamma$-TWIN$_w$ & $ {\cal L}_{rec} + \E[f(E_2 (x))] $\\
\hspace{3pt} $\sigma$-TWIN & 
${\cal L}_{rec}  +  \E [1-\cos(E_2(x),D_2(E_2(x))]  $\\
\end{tabular}
\end{table}
\begin{table*}[t]
\caption{ In-Alignment (IA) Training Data  \label{tbl:truth}}
\center\begin{tabular}{|p{0.45\textwidth}||p{0.4\textwidth}|}\hline
\hfill \sc  source    \hfill\null &\hfill  \sc target  \hfill\null \\ \hline
\small although the name suggests that they are located in the bernese oberland region of the canton of bern , portions of the bernese alps are in the adjacent cantons of valais , lucerne , obwalden , fribourg and vaud .  &\small 
although the name notes that they are in the bernese oberland region of the canton of bern , some of the
 bernese alps are found in the cantons of vaud , fribourg , valais , lucerne , uri , and nidwalden . \\ \hline
\small there he had one daughter , later baptized as mary ann fisher power , to ann -lrb- e -rrb- power . & \small 
there he had one daughter , later named mary ann fisher power , to ann -lrb- e -rrb- power .
\\ \hline 
\end{tabular}
\end{table*}
In this work, we look at a particular form of Wasserstein GAN (W-GAN)  which makes use of   the following loss function.
\begin{equation}
{\cal L} =  \E [f(x)]  - \E[ f(y)]  + \lambda G_p
\end{equation}
$\hat{x}$ is thought of as a sample arbitrarily chosen from a set of points between $x$ and $y$. $G_p$ is what is known as Gradient Penalty, which is meant to keep $\cal L$ within a certain range  \cite{lucic:2018}. $f$ can be any differentiable function that is 1-Lipschitz-continuous. $\lambda$ is a hyper-parameter to control the impact of $G_p$ on $\cal L$ (set to 1.0).  
\par  We make use of  the following as a discriminator for W-GAN:
 \begin{equation} \label{eqn:gp}
 f( H(x))   =  w^\top \text {\bf relu} (f_c ( H (x))))
 \end{equation}
$H (x) $ denotes a final state  we find ourselves in  after working through a sequence of words  that constitute  sentence $x$. $f_c$ represents a convolutional network consisting of a one-dimensional convolution layer and a max-pooling layer. {\bf relu} stands for rectified linear unit defined as $y = max (x, 0)$.  $w$ is a trainable weight matrix. 
\par JS-GAN uses  a discriminator that looks pretty much like  Eq.~\ref{eqn:gp}:
 \begin{equation}
 \begin{array}{lcl}
 D( H(x)) &  = & \text{\bf sigmoid}( f(H(x)) ), \\
\end{array}
 \end{equation}
 except that it has an additional activation function, intended  to make $D$'s  output compatible with the Jensen-Shannon loss (Eq.~\ref{eqn:js}). Other than that, $f$ and $D$ are identical. 
 
 \par Table~\ref{tbl:train-gans} gives a side-by-side comparison of JS- and W-GAN. Table~\ref{tbl:twin-loss} shows a loss function at play in each version  of the Twin Networks. 

\subsection{Flip-Flop Auto-Encoder (FFA)}

The problem with \cite{surya-etal-2019-unsupervised}, the latest recent bid to build a fully unsupervised text simplification, is that it had overlooked a particular relation  a simplified sentence  has with  a corresponding complex sentence: most of the time, it is a close copy of the source (Table~\ref{tbl:truth}).  Which motivates our idea of flip-flop auto-encoder (FFA).  FFA is like a regular translation model composed of an encoder and a decoder, converting the source to the target,  except that  it randomly flips itself into an auto-encoder  mode  where  the target  is switched from $y_k$ to $x_a$.   The decision to go for this particular arrangement was prompted by our  observation that using the translation model alone often resulted in mode collapse, the meaning of the original sentence  completely lost in translation.  By putting the model into an auto-encoder mode, we are hoping to see  the decoder retain  more of the semantic attributes  of the source sentence and generate a sentence that broadly looks like it.  As we  demonstrate later,  GAN is not enough to warrant the style transfer when  data are  not  properly aligned.

\renewcommand{\vec}[1]{\mathbf{#1}}
 
\begin{table*}[t]
\caption{ Out-of-Alignment (OOA) Training Data \label{tbl:train-samples}}
\center\begin{tabular}{|p{0.45\textwidth}||p{0.4\textwidth}|}\hline
\hfill \sc source  \hfill\null &\hfill  \sc target \hfill\null \\ \hline
\small six centuries later the king of kotte , veera parakramabahu viii ( 14771496 ) , had a network of canals constructed connecting outlying villages with colombo and negombo lagoon so that produce such as arecanuts , cloves , cardamom , pepper and cinnamon , could be more easily transported to the kingdoms main seaport at negombo .
&
\small 
at the top of the head there were five small holes , through which food would be ingested and waste products discharged . \\ \hline
\small in 1874 , he was nominated by the liberal-conservative convention as local candidate for the county of cumberland . &
\small 
he scored 107 goals in 429 league games in a 17-year career in the football league and scottish football league .\\ \hline
\end{tabular}
\end{table*}
\section{The Setup}
We build Twin Networks off of  Fairseq's multi-lingual translation module \cite{ott2019fairseq}. We assume that an encoder and a decoder we use are both LSTM based RNNs, which are shared across  two networks  we call UPPER and LOWER (Fig.~\ref{fig:base_model}).   We require that a sentence we feed as a target to LOWER   be   totally out of tune or alignment with its source as shown in Table~\ref{tbl:train-samples} (none of them remotely resembles simplifications of corresponding sentences).  Our goal  is to see whether Twin Networks are  able to learn to manage simplification,  given only  out-of-tune targets. We also look at effects of GAN, FFA, and  the similarity loss  on  performance.
 \par The training data come from a repository  \cite{surya-etal-2019-unsupervised} made available on GitHub\footnote{\url{https://github.com/subramanyamdvss/UnsupNTS.git} \label{fn:unts}}  (SOURCE, hereafter).  It has  2 million pairs of hard and easy sentences collected from Wikipedia, which are left unaligned. 

 \begin{table*}
 \caption{Key Numbers on Datasets. TOK: the average  (avg.) number of tokens per sentence. FRE:  avg. Flesch Reading Ease (more is better). FGL: avg. Flesch-Kincaid Grade Level (less is better). SIM: the avg.  (token based) cosine   similarity between HARD and EASY. SIZE: the number of instances in the relevant dataset.  \enquote*{OOA} is a short hand for \enquote*{Out of Alignment.} \enquote*{ALG} indicates whether source and target are  in  alignment. \label{tbl:stat}}
\hfill   \begin{subfigure}[h]{0.28\textwidth}
\hspace{-.0cm}
\begin{tabular}{ccc|}
& \multicolumn{2}{c|}{\sc source} \\
& \sc hard & \sc easy \\
TOK & 34.30 &18.89 \\
FRE & 32.91 & 83.95\\   
FGL & 17.37&  6.41\\\    
SIM & \multicolumn{2}{c|}{0.0032} \\
SIZE &\multicolumn{2}{c|}{2,000,000} \\
ALG &\multicolumn{2}{c|}{\sc no} \\
\end{tabular}
\end{subfigure} 
\hspace{-.6cm}
\begin{subfigure}[h]{0.19\textwidth}
\begin{tabular}{cc}
 \multicolumn{2}{c}{\sc ooa-train} \\
\sc hard & \sc easy \\
34.39& 18.44\\  
 32.15 & 85.43\\  
 17.65&  6.09\\     
 \multicolumn{2}{c}{0.0030} \\  
 \multicolumn{2}{c}{1,000,000} \\
 \multicolumn{2}{c}{\sc no} \\
\end{tabular}
\end{subfigure} 
\hspace{-.4cm}
 \begin{subfigure}[h]{0.14\textwidth}
\begin{tabular}{cc}
 \multicolumn{2}{c}{\sc ia-test} \\
 \sc hard & \sc easy \\
 22.61 & 22.15\\
 67.55 & 77.22\\
 9.51&  8.03\\
 \multicolumn{2}{c}{0.7767} \\
   \multicolumn{2}{c}{359} \\
  \multicolumn{2}{c}{\sc yes} \\
\end{tabular}
\end{subfigure} \hfill 
\hspace{-.0cm}
 \begin{subfigure}[h]{0.19\textwidth}
\begin{tabular}{|cc}
 \multicolumn{2}{|c}{\sc wikilarge} \\
 \sc hard & \sc easy \\
 25.11& 18.46\\
 63.79 & 74.79\\
 10.82&  7.71\\
 \multicolumn{2}{|c}{0.5731} \\
  \multicolumn{2}{|c}{296,402} \\
 \multicolumn{2}{|c}{\sc yes} \\
\end{tabular}
\end{subfigure} 
\hspace{-.4cm}
 \begin{subfigure}[h]{0.19\textwidth}
\begin{tabular}{|cc}
 \multicolumn{2}{|c}{\sc sscorpus} \\
 \sc hard & \sc easy \\  
 25.26& 17.95\\   
 62.26 & 74.52\\  
 10.87&  7.35\\
 \multicolumn{2}{|c}{0.6679} \\
  \multicolumn{2}{|c}{492,993} \\
  \multicolumn{2}{|c}{\sc yes} \\
\end{tabular}
\end{subfigure} 
\hfill \null
\end{table*}
Key statistics on  datasets  used for the experiment are given in Table~\ref{tbl:stat}.   Data for training were created using the first 1,000,000 pairs of SOURCE (call it OOA-TRAIN).   IA-TEST represents  a popular bench mark test set created  by \cite{zhang-lapata-2017-sentence}. Each test instance comes with  eight human made references for each source sentence. HARD refers to sentences that are considered to be \enquote*{hard,} EASY to  sentences that are \enquote*{easy.}  That HARD and EASY are indeed what they are supposed to be is vindicated by statistics given in   Table~\ref{tbl:stat}.
We included for reference,  WikiLarge \cite{zhang-lapata-2017-sentence},  a popular Wikipedia based corpus widely used in supervised simplification, composed of  sentence-by-sentence alignments of hard and  easy sentences, which were manually verified. We used as  a development set one prepared by \cite{zhang-lapata-2017-sentence}, made up of 992 in-alignment pairs. SSCORPUS is another dataset created by \cite{Tomoyuki_Kajiwara2018} from Wikipedia for text simplification.  Like WikiLarge, its source and target sentences are put  into alignment, but unlike WikiLarge,  they are both sourced from Wikipedia and constructed with no human intervention, relying instead on some similarity metric. 
 \par There are some additional points about Table~\ref{tbl:stat} which are worth mentioning.   The similarity between HARD and EASY in IA-TEST (which is at 0.57) is much closer to WikiLarge (0.77)  than what we have in OOA-TRAIN and  SOURCE (0.0032 and 0.0031, respectively),  supporting the view  that a simplification and its source  are quite alike in appearance.\footnote{The similarity was measured by applying cosine over word tokens in sentences, which exclude stop words. The script for the similarity comes from \url{https://www.scipy.org/}.}  The closeness of IA-TEST to WikiLarge in  FRE and FGL suggests   that both were  derived from the same source \cite{zhang-lapata-2017-sentence}.  OOA-TRAIN (and SOURCE)  however, is a completely different creature. Consider its HARD portion.  It has  an FRE value  almost half that of WikiLarge and  an FGL  nearly twice as large. In addition, the similarity between HARD and EASY in OOA-TRAIN is effectively zero. 
 The question of  whether we can leverage OOA-TRAIN, to arrive at a model that produces decent simplifications, in the face of these obstacles is  discussed  in the section below. %
 \begin{table}[t]
\caption{Main Results with {\sc ooa train}.  We set the flip rate at 0.2 for  the Twin Networks except $\gamma$-{\sc twin}$_j^0$ whose rate was set to 0.0.  By setting the flip rate at 0.2, we  are sending  the model into  an auto-encoder mode 20\% of the time. $\gamma$-{\sc twin}$_j$ and $\gamma$-{\sc twin}$_j^0$ are identical apart from the flip rate. The upward arrow means higher is better, the downward arrow lower is better. \label{tbl:main-results} }
\hspace{-7mm}
\begin{tabular}{lccccc}
& \sc bleu$\uparrow$  & \sc sari$\uparrow$ & \sc fgl$\downarrow$ & \sc fre$\uparrow$ & \sc diff$\uparrow$\\
\sc $\beta$-twin&  \bf 0.970 & 0.292 & 13.32 & 56.62 & 0.31\\
\sc $\gamma$-twin$_j$ &  0.880 & \bf 0.344 &  \bf  8.11& 70.91 & \bf 4.75\\
\sc $\gamma$-twin$_j^0$ & 0.508 & 0.337 & 8.38 & \bf 89.59 & 0.45 \\
\sc $\gamma$-twin$_w$  & 0.919 & 0.316 & 11.26 & 63.08 & 1.89\\
\sc $\sigma$-twin & 0.955 & 0.305 & 12.32 & 59.56 & 0.77\\ \hline
\sc unts& 0.356  & 0.307 &  7.54 & 84.09 &  9.26 \\
\sc sour & 0.993 & 0.279 & 9.51 & 67.56 &  0.00 \\ 

\end{tabular}
\end{table}

 \section{Results and Discussion \label{sec:results}}

For the experiment,  we worked with four metrics, BLEU, SARI, FGL, FRE, and DIFF.\footnote {Scripts for BLEU and SARI come from \url{ https://github.com/cocoxu/simplification}. FRE is from \url{https://pypi.org/project/readability}.  FGL  is from \url{https://github.com/nltk}.}  
 BLEU \cite{papineni-etal-2002-bleu} is a metric widely used as a standard yard stick in the machine translation community and beyond.
SARI \cite{xu-etal-2016-optimizing} is a recent addition, which essentially looks at  the quality of simplification from three perspectives: how different the output is from the source; how well the output retains words that should not be deleted; and how successful the output is in getting rid of words that do not appear in the reference.  
One caveat is that  scoring better with SARI (higher is better) does not necessarily mean that  we have a more readable or grammatical output. It just tells you how different the output is from the source in a way that agrees with the reference. DIFF, which measures the difference in length (word count) between the source and the output, i.e. {\sl length\_of (source) - length\_of(output)}, was put in here to check whether simplification results in a shorter sentence,  which we want to see happen. FGL (Flesch-Kincaid Grade Level)  and FRE (Flesch Reading Ease) are widely accepted metrics for the readability.

\par Table~\ref{tbl:main-results} looks at  main results with the Twin Networks trained on OOA data.  Along with Twin Networks, we are also showing the current state of the art, {UNTS} \cite{surya-etal-2019-unsupervised},   which we trained on OOA-TRAIN.\footnote{For the test, we  used a model that came about at the 13,000-th training step  in accordance with the setting given in the code at \url{https://github.com/subramanyamdvss/UnsupNTS}. The batch size was set at 12  due to the limitation of the GPU memory at our disposal. Everything else was set to defaults ({mgan} on and { backtranslation} off).}  {SOUR} represents a case where we treat the source as output, which would give us an idea of what happens if no change is made to the source. Given the results, we can argue with confidence that   $\gamma$-{TWIN}$_j$  is superior to  {UNTS}. While maintaining the high degree of BLEU and SARI, it managed to shorten the sentence by 4.75 words on average.  The fact that  there is not much difference in DIFF and in SARI for  other {TWIN}  systems  suggests that their simplifications would look  pretty much like source sentences.  Finally, $\gamma$-{TWIN}$_j^0$ shows what happens if we disengage  the flip-flop capability  for $\gamma$-{TWIN}$_j$. Its resulting degradation in BLEU clearly demonstrates the efficacy of FFA.

\par Another important point to make is that $\gamma$-{\sc twin}$_w$ proved not as successful as $\gamma$-{\sc twin}$_j$,  implying that Wasserstein-GAN may not be a good choice when it comes to moving the readability attribute from a short to long sentence.  Although the reason is not immediately clear,  it is likely that it may have been distracted by factors not relating to readability, as its focus is more on the similarity of representations produced by encoders than on whether they are readable. 

\par At this point, we look at where we stand in comparison to the prior work that addressed simplification with no explicit supervision beyond UNTS.  We focus on two approaches. One is   \cite{glavavs-vstajner:2015:ACL-IJCNLP} who introduced an approach called  {LIGHT-LS} ({LLS}) which  is based on an idea that one would be able to get a sentence simplification  by replacing some words with those with a similar meaning but with more informative content, which is determined using the likes of Word2Vec and TFIDF, both of which are pre-computed. Simplification is achieved  by simply looking up a weight dictionary built in advance. 

\par  Another avenue, explored by \cite{Tomoyuki_Kajiwara2018}, focuses on constructing a corpus that serves as a replacement for a true parallel corpus rather than building  a model that works in the absence of supervision. The author proposed a particular similarity metric called MAS, to find targets for source sentences, both from Wikipedia.  An important finding by the work is that a target needs not to  be a strict simplification of the source  in order for the training to work. The study has shown that a phrase based statistical machine translation model (PB-SMT) \cite{koehn:etal:2007:moses}\footnote{\url{http://www.statmt.org/moses/}} trained on a synthetic corpus worked just as well as one  trained on an  \enquote*{authentic} in-alignment corpus like WikiLarge  \cite{zhang-lapata-2017-sentence},  which  draws on Simple Wikipedia, a simplified version of Wikipedia.\footnote{\url{https://simple.wikipedia.org/wiki/Main_Page}} The process, however,  contains a crucial step, one that  requires us to go through every possible pair of sentences   to arrive at the corpus, which  is exactly what we have sought to  avoid in this work.  {SSCORPUS}, a corpus the author published,\footnote{\url{https://github.com/tmu-nlp/sscorpus.git}}   had a SIM value of 0.67 which far exceeds that of WikiLarge,  which stands at 0.57 (Table~\ref{tbl:stat}).  In addition,  it is striking  that the minted corpus and WikiLarge look quite alike in readability metrics: hard and easy sentences in the former are of about the same length as those in the latter, respectively, and the former's  scores on FRE and FGL are  largely comparable to those for WikiLarge. 

\par Table~\ref{tbl:sup-free-systems} shows  how LLS and PB-SMT fared in the current setup. We trained PB-SMT on  SSCORPUS, and tested on {\sc IA-TEST} (details in Table~\ref{tbl:stat}). We ran LLS  on {IA-TEST} and collected whatever it produced. The table also shows   $\gamma$-TWIN$_j$ alongside.  We see PB-SMT and LLS exhibiting a comparable performance.  In either case, the length of output remains close to that of the source.  $\gamma$-TWIN$_j$  is doing almost as good as the other two.  What sets it apart is the length of outputs, which is about a fourth that of sentences produced by the two. 

\begin{table}
\caption{Previous \enquote{Supervision-free} Systems \label{tbl:sup-free-systems}}
\hspace{-6mm}
\begin{tabular}{lccccc}
& \sc bleu$\uparrow$  & \sc sari$\uparrow$ & \sc fgl$\downarrow$ & \sc fre$\uparrow$ & \sc diff$\uparrow$\\
\sc pb-smt  &  0.896 & 0.352 &  9.15& 70.06&0.11\\
\sc lls   & 0.738& 0.349  &  8.93&  71.62& 0.19 \\  \hline
\sc $\gamma$-twin$_j$ &  0.880 & 0.344 &  8.11& 70.91 & 4.75\\
\end{tabular}
\end{table}

\begin{table}[t]
\caption{Supervised Systems trained on WikiLarge.\label{tbl:sup-results}}
\hspace{-6mm}\begin{tabular}{cccccc}
& \sc bleu$\uparrow$  & \sc sari$\uparrow$ & \sc fgl$\downarrow$ & \sc fre$\uparrow$ & \sc diff$\uparrow$\\
\sc dress & 0.772 & 0.371 & 6.80 & 75.75 & 6.46 \\
\sc dress-ls & 0.801& 0.373  & 6.92 & 75.30  & 6.21 \\  
\sc fconv & 0.822 & 0.355  & 7.12 & 73.56& 6.40 \\  \hline 
\sc human & 0.702 & 0.411 & 8.03 & 77.22 & 0.45\\
\end{tabular}
\end{table}

\begin{table*}[t]
\caption{$\gamma$-TWIN$_j$  vs. Prior Supervision-free Systems\label{tbl:generated}}
\begin{center}
\begin{tabular}{|p{0.1\textwidth}||p{0.75\textwidth}|} \hline
{\sc\bf  source} & \small in architectural decoration small pieces of colored and iridescent shell have been used to create mosaics and inlays , which have been used to decorate walls , furniture and boxes . \\ \hline
$\gamma$-{\sc twin$_j$} &  \small 
in architectural decoration small pieces of colored and iridescent shell have been used to create mosaics and inlays .\\\hline
{\sc pb-smt} & \small 
in architectural decoration , small pieces of colored and iridescent shell have been used to create various and imitating , which have been used to decorate walls , furniture and boxes . \\  \hline
{\sc unts } & \small in small small small style of white used to create to sand and inlays , which have been used to various walls , and stones . \\ \hline
{\sc lls} & \small  in {\color{blue} art} decoration small pieces of bright and metallic shell have been used to create paintings and inlays , which have been used to painted walls , wood and filled .\\ \hline
{\bf target} & \small small pieces of colored shell and iridescent shell have been used to create mosaics and inlays which have been used to decorate larger items such as boxes and furniture . \\ \hline
\hline {\bf  source} & \small
aside from this , cameron has often worked in christian-themed productions , among them the post-rapture {films} left behind : the movie , left behind ii : tribulation force , and left behind : world at war , in which he plays cameron " buck " williams . \\ \hline
$\gamma$-{\sc twin$_j$} &  \small 
aside from this , cameron has often worked in christian-themed productions . \\ \hline
{\sc pb-smt } & \small cameron has often worked in christianity-related movies , among them the post-rapture {\color{blue} movies} left behind : the movie , left behind ii : establish force , and left behind : world at war , in which he plays cameron `` buck " " williams . \\ \hline
\sc unts & \small  apart from this , , he has often worked in christian-themed production , among them post-rapture , post-rapture left behind the movie , left behind ii behind tribulation force , and left left behind world at war . \\ \hline
{\sc lls} & \small aside from this , cameron has often worked in christian-themed directed , among them the post-rapture directed left behind : the film , left behind ii : tribulation force , and left behind : world at war , in which he plays cameron " buck " smith .\\ \hline
{\bf target} & \small cameron has also often worked in christianity-related movies , among them the post-rapture movies left behind : the movie , left behind ii : tribulation force , and left behind : world at war , in which he plays cameron " buck " williams .\\ \hline
%
\end{tabular}
\end{center}
\end{table*}
\par   Table~\ref{tbl:sup-free-systems}, however,  is not  so clear about  what qualitative differences there are if any, among the systems, with their performances coming  in  a close range of  one another. A glance at some of the simplifications by the systems however, reveals substantive  differences that exist in quality, which are shown in Table~\ref{tbl:generated}. What immediately comes to our attention  is how short  the sentences  produced by $\gamma$-TWIN$_j$ are:   while maintaining grammaticality,  it tends to split off  the source sentence   at a major constituent break. In PB-SMT and LLS, changes  made to the source are confined to minor lexical replacements: \enquote*{art} for \enquote*{architectural} in LLS and \enquote*{movies} for \enquote*{films} in PB-SMT (all shown in blue).

\par To set our results in a broader perspective,  let us  look at how far  we  are from those that are fully supervised. Table~\ref{tbl:sup-results} gives  performance of  supervised systems trained on WikiLarge and tested on IA-TEST.  All the systems (except  HUMAN)  come from the published literature. HUMAN represents the ground truth.   FCONV is  a translation model based on a convolutional neural network \cite{ott2019fairseq}.\footnote{\url{https://github.com/pytorch/fairseq}} 
 DRESS \cite{zhang-lapata-2017-sentence}  is  a cookie-cutter sequence to sequence model using  REINFORCE \cite{williams:1992} as a sole objective.  DRESS-LS is a DRESS with an added reconstruction loss.   
  \begin{table*}
\caption{Amazon MTurk Ratings \label{tbl:human_ratings}} 
\center\begin{tabular}{cccccc}
& fluency & grammar &  readability  & meaning & length (\# of tokens) \\
source & 4.46 & 4.42 & 4.45 & - & 27.02 \\
output  &  3.87 &  3.88 & 4.01& 3.60 & 20.93\\
\end{tabular}
\end{table*}
 As is clear from the table, a most visible difference between fully supervised systems  (FSSs) and ours lies in how diverse the  outputs are.  Results on SARI and DIFF make a persuasive case  that simplifications from FSSs are far more varied  in a way consistent with human judgments, with their length markedly shorter than what we were able to achieve with $\gamma$-TWIN$_j$. 
  \par With these observations in hand, we may conclude that 
 the lack of alignment makes  it  extremely difficult for the system  to discover patterns in syntax and word associations that could be tapped into to diversify a sentence  it generates.

\section{Human Evaluation}
We ran a  survey to look at how humans may react o texts generated by $\gamma$-TWIN$_j$, comparing what we got before and after the transformation.  We made use of the Amazon Mechanical Turk (AMT) platform. Each of the HITs (Human Intelligence Task) consisted of a pair of sentences, one from the source (IA-TEST HARD) and the other from what was generated by $\gamma$-TWIN$_j$, which corresponds to the former. We asked each of the Workers (we had 720 of them)  to rate the pair on a scale of 1 (poor) to 5 (excellent), in terms of grammaticality, readability, fluency, and how well meaning is preserved under transformation. For this task, we focused on those outputs of $\gamma$-TWIN$_j$ that are shorter than and not identical to source sentences.\footnote{Translation onto itself was a phenomenon that we often saw with $\gamma$-TWIN$_j$, which is largely due to FFA.  } The question we were interested in was whether the length has any effect on readability.  We put out 120 test pairs, each of which was assigned to 6 Workers who self-reported that their education level is  college or higher,  and they  spoke English as a native language. We restricted participation to English speaking regions, US, CA, UK, AU, and NZ.  \par The result is shown in Table~\ref{tbl:human_ratings}. 
While  simplifications we produced are not significantly worse off than their originals,  their poor performance  on readability is something of a concern. It suggests that the length does not play much role in the business of simplification. The focus of this work is more on finding whether stylistic traits can be transferred across sentences without relying on the ground truth than on how much they contribute to simplification.  Therefore, the result from human evaluation will not discredit the claim we made in Section~\ref{sec:results}.  Nonetheless, we may have to live with  the possibility that  the length is not as relevant to simplification as we thought it was.


\section{Conclusion}
It turned out that the answer to the question we posed at the start of the paper, of whether we can do simplification without out of line data was yes. 

\par Additionally, we found: (1) $\gamma$-TWIN$_j$, Twin Networks with JS-GAN regulating encoder outputs, outperformed the previous best, UNTS; (2) JS-GAN proved to be a crucial piece in boosting performance,   as a comparison with $\gamma$-TWIN$_w$  would confirm; (3)   a drop in BLEU seen by  $\gamma$-TWIN$_j^0$ (Table~\ref{tbl:main-results}) demonstrates that  Flip-Flop Auto-Enocder (FFA), is a critical component to  ensuring the grammatical integrity of outputs;  (4)  while performances among supervision-free systems (PB-SMT, LLS, $\gamma$-TWIN$_j$) are hard to distinguish in   BLEU and SARI,  a close look at examples  revealed  significant qualitative differences that exist (Table~\ref{tbl:generated}); (5) we were unable  to attain structural and lexical diversity on par with supervised systems (Table~\ref{tbl:sup-results}), which we view as a fundamental limitation of  the current setup; (6) evaluation with humans found that while we were able to move some stylistic features (in particular, length) across sentences using the Twin Network model, it did not result in an improvement in readability with Amazon Mechanical Turk, a revelation that demands that we rethink the way we approach simplification as we go forward with the research.
\bibliographystyle{acl_natbib}
\bibliography{ling1-u_emnlp}
\end{document}